\documentclass[lettersize,journal]{IEEEtran}
\usepackage{amsmath,amsfonts}
\usepackage{algorithmic}
\usepackage{algorithm}
\usepackage{array}
\usepackage[caption=false,font=normalsize,labelfont=sf,textfont=sf]{subfig}
\usepackage{textcomp}
\usepackage{stfloats}
\usepackage{url}
\usepackage{verbatim}
\usepackage{graphicx}
\usepackage{cite}
\usepackage{multirow}
\usepackage{xcolor}
\usepackage{booktabs}
\usepackage{hyperref}
\begin{document}

% \title{Global and Local Differences of Frequency for High Dynamic Range Image Quality Assessment}
% \title{High Dynamic Range Image Quality Assessment via Global and Local Differences of Frequency}
% \title{High Dynamic Range Image Quality Assessment via Global and Local Frequency Character}
% \title{High Dynamic Range Image Quality Assessment via Frequency Differences }
\title{High Dynamic Range Image Quality Assessment Based on Frequency Disparity}
% \title{LGFM: High Dynamic Range Image Quality Assessment Via Global and Local Frequency feature-based Model}
% \title{Visual Frequency Feature Decomposition for High Dynamic Range Image Quality Assessment}
% \title{High Dynamic Range Image Quality Assessment in frequency domain}

\author{Yue~Liu,~\IEEEmembership{Student Member,~IEEE}, Zhangkai Ni,~\IEEEmembership{Member,~IEEE}, Shiqi Wang,~\IEEEmembership{Senior Member,~IEEE}, Hanli~Wang,~\IEEEmembership{Senior Member,~IEEE,} and Sam Kwong,~\IEEEmembership{Fellow,~IEEE}
        % <-this % stops a space
        
% \thanks{This work was supported in part by the Key Project of Science and Technology Innovation 2030 supported by the Ministry of Science and Technology of China under Grant 2018AAA0101301, in part by the Natural Science Foundation of China under Grant 61772344, and Grant 61672443, in part by the Hong Kong Research Grants Council (RGC) General Research Funds under Grant 9042816 (CityU 11209819), and under Grant 9042957 (CityU 11203220), in part by the Hong Kong Research Grants Council (RGC) Early Career Scheme under Grant 9048122 (CityU 21211018), and in part by the Hong Kong Innovation and Technology Commission (InnoHK Project CIMDA).}

\thanks{Yue Liu, Shiqi Wang and Sam Kwong are with the Department of Computer Science, City University of Hong Kong.  (e-mail:yliu724-c@my.cityu.edu.hk;  shiqwang@cityu.edu.hk; cssamk@cityu.edu.hk).}% <-this % stops a space
%thanks{Zhangkai Ni is with the Department of Computer Science & Technology and Key Laboratory of Embedded System and Service Computing (Ministry of Education) Tongji University, Shanghai 200092, P. R. China (e-mail: eezkni@gmail.com).}% <-this % stops a space
\thanks{Zhangkai Ni and Hanli Wang are with the Department of Computer Science and Technology, Key Laboratory of Embedded System and Service Computing (Ministry of Education), and Shanghai Institute of Intelligent Science and Technology, Tongji University, Shanghai 200092, P. R. China (e-mail: eezkni@gmail.com; hanliwang@tongji.edu.cn).}% <-this % stops a space
%\thanks{is with the Department of Computer Science, City University of Hong Kong, Hong Kong 999077, and also with the City University of Hong Kong Shenzhen Research Institute, Shenzhen 518057, China (e-mail: ).}
}



\maketitle

\begin{abstract}
In this paper, a novel and effective image quality assessment (IQA) algorithm based on frequency disparity for high dynamic range (HDR) images is proposed, termed as local-global frequency feature-based model (LGFM).
Motivated by the assumption that the human visual system is highly adapted for extracting structural information and partial frequencies when perceiving the visual scene, the Gabor and the Butterworth filters are applied to the luminance of the HDR image to extract local and global frequency features, respectively. The similarity measurement and feature pooling are sequentially performed on the frequency features to obtain the predicted quality score. The experiments evaluated on four widely used benchmarks demonstrate that the proposed LGFM can provide a higher consistency with the subjective perception compared with the state-of-the-art HDR IQA methods. Our code is available at: \url{https://github.com/eezkni/LGFM}.
% In this paper, a novel and accurate full-reference image quality assessment (IQA) model for the high dynamic range (HDR) image is proposed by using the frequency differences, called the local and global frequency feature-based model (LGFM). Motivated by the observation that the human visual system (HVS) is sensitive to the edge information and part of the spatial frequency, the proposed model first extract the local and global frequency features. Afterwards, the similarity measurement and pooling strategy are conducted on the feature maps to generate the similarity score of the local and global frequency features, which are combined to obtain the final HDR quality score. The experimental results have shown that the proposed LGFM provides higher consistency with the HVS perception on the evaluation of the HDR image quality compared with the state-of-the-art IQA metrics.
\end{abstract}


\section{Introduction}
\IEEEPARstart{W}{ith} the rapid development of imaging technology and the growing demand for immersive experiences, the \textit{high dynamic range} (HDR) images are increasingly indispensable due to the realistic experiences they can provide, which can significantly contribute to the development of TV and photography industry. 
Compared with the 8-bit \textit{low dynamic range} (LDR) images, HDR images are linearly related to the physical luminance in the scene and can record more structural details by using 16-32 bit floating point values~\cite{dufaux2016high}. 
Essentially speaking, there are two major differences between HDR and LDR images, 1) the data distribution of HDR images is much broader than that of LDR images; 2) more detailed structures can be preserved in HDR images. 
Therefore, most \textit{image quality assessment} (IQA) methods designed for LDR images are not suitable for direct use in assessing the quality of HDR images, which makes it crucial for developing effective IQA models for HDR images.

\begin{figure*}
\begin{center}
  \includegraphics[width=0.85\linewidth]{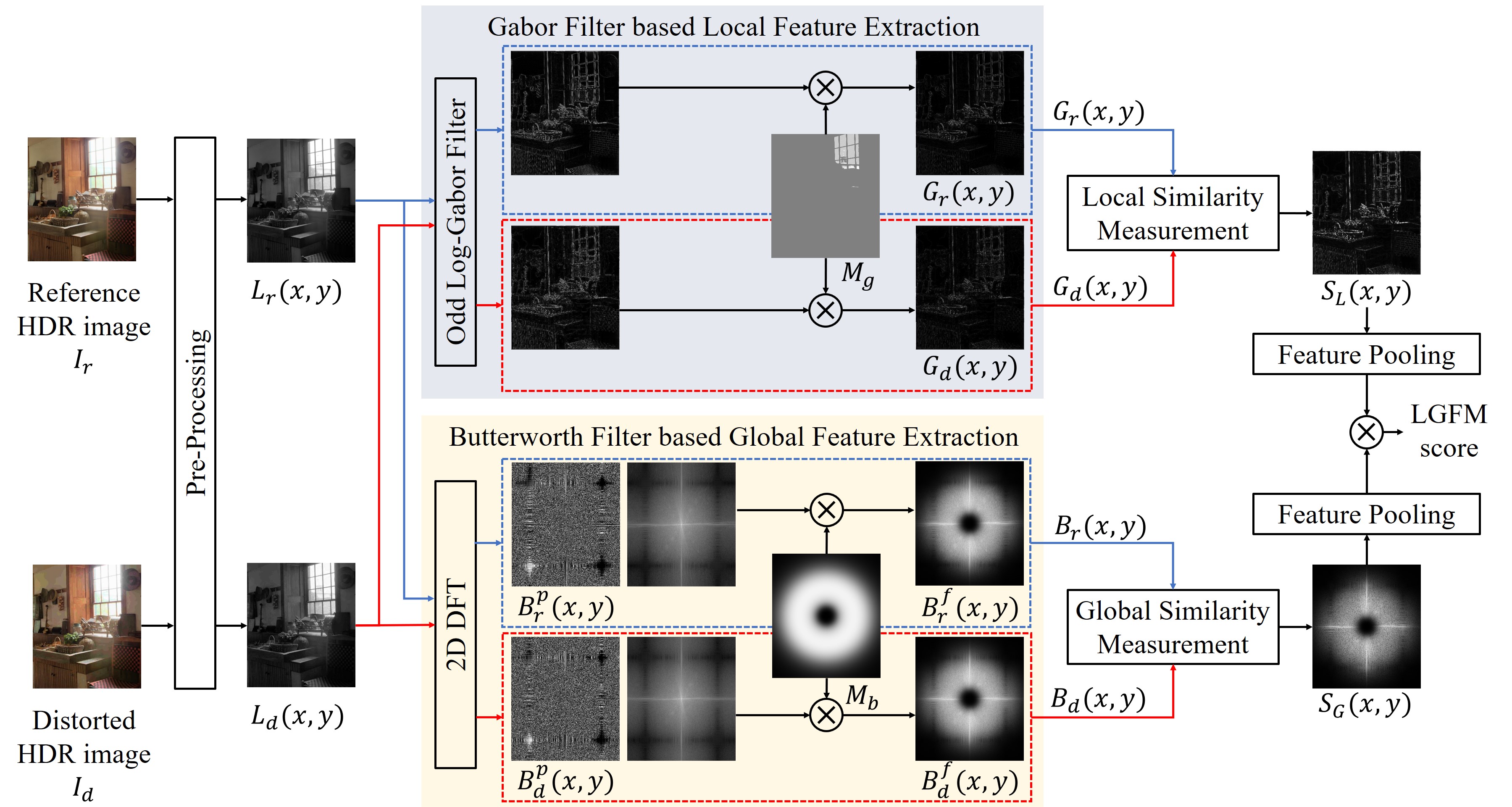}
\end{center}
\vspace{-10pt}
  \caption{
  The flowchart of the proposed \textit{Local and Global Frequency feature-based Model} (LGFM) for HDR IQA. 
  Firstly, The reference and distorted HDR images, $I_r$ and $I_d$, are converted to the luminance perceptual space, $L_r$ and $L_d$, in the pre-processing stage. 
  Subsequently, an odd log-Gabor filter is designed to extract local frequency features, $G_r$ and $G_d$, with a spatial mask $M_g$ used to provide higher weights to the over-exposed region.
  Meanwhile, the Butterworth filter is used to simulate the contrast sensitivity function to highlight the sensitive frequency interval in HVS by the generated mask $M_b$.
%   The generated mask $M_b$ can highlight the sensitive frequency interval in HVS. 
  The global frequency features of the reference and distorted HDR images, $B_r$ and $B_d$, are consisted of the corresponding phase maps and the weighted frequency maps. 
  Finally, the local and global similarities are measured, followed by the feature pooling strategy to predict the quality score.
%   The framework of the proposed \textit{Local and global frequency feature-based model} (LGFM) for HDR IQA. 
%   Firstly, The reference and distorted HDR images are converted to the luminance maps by preprocessing, $L_r(x,y)$ and $L_d(x,y)$. 
%   Subsequently, an odd log-Gabor filter is designed to extract local frequency features $G_r(x,y)$ and $G_d(x,y)$, with a spatial mask $M_g$ used to provide higher weights to the over-exposed region.
%   In addition, the Butterworth filter is used to simulate the contrast sensitivity function. 
%   The generated mask $M_b$ can highlight the sensitive frequency interval in HVS. 
%   The global frequency features of the reference and distorted HDR images $B_r(x,y)$ and $B_d(x,y)$ consist of the corresponding phase maps and the weighted frequency maps. 
%   The local and global similarities are measured, followed by the pooling strategy to generate the final quality score.
  }
\label{fig:blocks}
\vspace{-10pt}
\end{figure*}

%In order to simulate the real perception of human beings, 
The IQA models aim to objectively evaluate the image quality to align with the \textit{human visual system} (HVS). 
As a fundamental problem in the field of image processing, IQA models have been widely used to optimize the performance of various learning-based vision tasks and to improve image encoding capability by increasing compression ratio while preserving the original image quality. 
According to the amount of available information of the reference image, existing IQA models can be roughly divided into three categories, \textit{full-reference} (FR)~\cite{fu2018screen}~\cite{tian2020light}, \textit{reduced-reference} (RR)~\cite{wang2016reduced}~\cite{huang2020perceptual}, and \textit{no-reference} (NR)~\cite{ou2021novel}~\cite{pan2022vcrnet}.
% Existing IQA models can be roughly divided into three categories, \textit{full-reference} (FR)~\cite{fu2018screen}~\cite{tian2020light} where the reference image is accessible, \textit{reduced-reference} (RR)~\cite{wang2016reduced}~\cite{huang2020perceptual} where partial character of the reference image is available, and \textit{no-reference} (NR)~\cite{ou2021novel}~\cite{pan2022vcrnet} where the reference image is not provided. 
In this paper, we are committed to proposing a novel and effective FR IQA model for HDR images.

The most widely used objective IQA measure for HDR images is the \textit{HDR visual difference predictor} (HDR-VDP) proposed by Mantiuk \textit{et al.}~\cite{mantiuk2011hdr}, the latest release of which is HDR-VDP-3~\cite{narwaria2015hdr}. 
They model the HVS by taking into account optical and retinal pathways in the human eye and evaluating visible differences between images. The \textit{dynamic range independent metric} (DRIM) proposed by Aydin \textit{et al.}~\cite{aydin2008dynamic} models the HVS and presents the differences with three distortion maps, which is difficult for subsequent analysis due to the lack of a pooling strategy. Currently, there is a metric specifically designed for the HDR video content, namely \textit{HDR video quality metric} (HDR-VQM) proposed by Narwaria \textit{et al.}~\cite{narwaria2015hdrvqm}. The input HDR video sequences are first projected to the perceptual domain, then the Gabor filter is applied to extract the frequency features, which are used to calculate the subband errors. The final quality score is obtained through spatial and temporal pooling. Nevertheless, both the HDR-VDP-3 and the HDR-VQM require additional viewing information such as display parameters and viewing distance.

Due to the lack of IQA measures specifically designed for HDR images, various objective LDR IQA models are also used to evaluate the quality of HDR images. However, the data distribution of the HDR image is quite different from that of the LDR image. 
% In LDR imaging, traditional \textit{camera response functions} (CRFs) are usually adopted to simulate the non-linear luminance perception of the HVS due to the limitation of the storage space (8 bit) and the maximum luminance value of displays ($100cd/m^2$). 
% In contrast, HDR imaging can store luminance values that are linear to the natural light (up to $5000cd/m^2$)~\cite{reinhard2010high}. 
Since traditional CRFs are designed for dim LDR displays, Aydin \textit{et al.}~\cite{aydin2008extending} proposes the \textit{perceptual uniform} (PU) space to map a wide range of luminance to a perceptual range that is consistent with the HVS. 
Therefore, the quality of encoded HDR images can be evaluated with LDR measures, such as \textit{peak signal-to-noise ratio} (PSNR), \textit{structural similarity} (SSIM)~\cite{wang2004image}, \textit{gradient magnitude similarity deviation} (GMSD)~\cite{xue2013gradient}, \textit{visual information fidelity} (VIF)~\cite{sheikh2005information}, and \textit{feature similarity} (FSIM)~\cite{zhang2011fsim}. 
In analogous to the PU function, the \textit{perceptual quantizer} (PQ) proposed in~\cite{standard2014high} can be used in the same way. 
\textit{HDR-Combination of Quality Metrics} (HDR-CQM) proposed by Choudhury~\cite{choudhury2020robust} improves the performance of HDR IQA by taking advantage of various HDR and LDR IQA metrics. 
Recently, Mikhailiuk \textit{et al.}~\cite{mikhailiuk2021consolidated} construct a \textit{Unified Photometric Image Quality dataset} (UPIQ) by aligning and merging two HDR and two LDR image datasets for IQA, based on which the PU-PieAPP model is trained for HDR IQA. 

% Although great efforts have been made on HDR IQA and achieve the state-of-the-art results, 
In this paper, a novel full-reference HDR IQA metric is constructed by modeling the HVS-sensitive information with local and global frequency characters to further improve the assessment accuracy. 
The proposed \textit{Local and Global Frequency feature-based Model} (LGFM) is mainly inspired by two observations. First, the local frequency feature can well represent the texture details perceived by human eyes. Second, the global frequency feature can be used to characterize the sensitive frequency interval of the HVS. 
In the proposed method, the reference and distorted HDR images are first transferred to the perceptual space using the PU coding. 
Thereafter, local and global frequency feature maps are extracted by specifically designed Gabor filtering and Butterworth filtering, respectively. 
The similarity measurements and pooling strategies are performed on the feature maps to generate similarity scores of local and global frequency features, respectively. The final prediction quality score is obtained by multiplying these two scores.
% which are combined to obtain the final HDR quality score. 
% The experimental results have shown that the proposed LGFM provides higher consistency with the HVS perception on the evaluation of the HDR image quality compared with the state-of-the-art IQA metrics.
Abundant experiments demonstrate the superiority of the proposed LGFM in the evaluation of the HDR image quality compared with the state-of-the-art IQA algorithms. 

% The remainder of the paper is organized as follows. Section~\ref{sec:proposed} introduces the proposed LGFM in detail. Section~\ref{sec:results} presents extensive experimental results and discussions. Section~\ref{sec:conclusion} finally draws the conclusion.
%The organization of the remainder of the paper is as follows. Section~\ref{sec:proposed} describes detail structure of the proposed LGFM. Section~\ref{sec:results} represents extensive experimental results. Section~\ref{sec:conclusion} finally draws the conclusion.

% \section{Proposed Local and global frequency feature-based model for HDR images}
\section{Proposed Local and global frequency feature-based model}
\label{sec:proposed}

The framework of the proposed LGFM is illustrated in Fig.~\ref{fig:blocks}, which consists of four processing stages. 
Firstly, the reference and distorted HDR image, $I_r$ and $I_d$, are converted to the perceptual space in the pre-processing stage. 
The corresponding luminance maps, $L_r(x,y)$ and $L_d(x,y)$, are obtained from the linear luminance space by the PU coding, where $(x, y)$ denotes the pixel coordinate in the image.
In the second stage, the Gabor filter and Butterworth filter are used to extract the local frequency features ($G_r(x,y)$, $G_d(x,y)$) and global frequency features ($B_r(x,y)$, $B_d(x,y)$), respectively.
In the third stage, the computed frequency features from the reference and distorted HDR images are compared separately to yield the local and global similarity maps.
Finally, the two computed similarity maps are combined to generate the predicted quality score using the proposed feature pooling strategies.

% In the pre-processing stage, the reference HDR image $I_r$ and distorted HDR image $I_d$ are converted to the perceptual space, i.e., $L_r(x,y)$ and $L_d(x,y)$ from the linear luminance space by the PU coding [], where $(x, y)$ denotes the pixel coordinate in the image. 
% In the feature extraction stage, the Gabor filter and Butterworth filter are utilized to extract the local and global frequency features, $G_r(x,y)$, $G_d(x,y)$, $B_r(x,y)$ and $B_d(x,y)$, respectively. Afterward, the local and global similarity map, $S_l(x,y)$ and $S_g(x,y)$ are generated by combing and measuring the similarity between the reference and distorted frequency features. Finally, the similarity maps are weighted and combined to obtain the final quality score.

% \vspace{-10pt}
\subsection{Gabor Filter-based Local Feature Extraction}
The local frequency feature can be used to extract abundant structural and edge information, where the Gabor filter is greatly consistent with the response of the HVS
% , which is highly sensitive to edge information
~\cite{wang2004image, ni2017esim, ni2018gabor}. Therefore, we adopt the Gabor filter for local frequency feature extraction.
% the Gabor filter is adopted to extract local frequency features of HDR images.
% Since the Gabor filter is greatly consistent with the response of the HVS and the HVS is highly sensitive to edge information~\cite{wang2004image, ni2017esim, ni2018gabor}. 
% Typical HDR images contain abundant structural and edge information, which motivates us to exploit the Gabor filter to extract local structure features of HDR images~\cite{ni2018gabor}.
The Gabor filter-based features of reference and distorted HDR images are denoted as $G_r(x,y)$ and $G_d(x,y)$, respectively. 
% That is,
% \begin{equation}
% \begin{aligned}
% G_r(x,y) = \{G_r^g(x,y), G_r^l(x,y)\};\\
% G_d(x,y) = \{G_d^g(x,y), G_d^l(x,y)\},
% \end{aligned}
% \end{equation}
% where $\left((G_r^g(x,y), G_d^g(x,y)\right)$ are the global edge maps of reference and distorted HDR images. Similarly, $\left((G_r^l(x,y), G_d^l(x,y)\right)$ are the local edge maps of reference and distorted HDR images.

Inspired by work\cite{ni2018gabor}, the local edge information is extracted using the odd log-Gabor filter, which can well represent the high-frequency component of nature images, 
\begin{equation}
\begin{aligned}
& G_{o}(x,y) = \\
& \frac{1}{2\pi\sigma_x\sigma_y}\text{exp}\left\{ \frac{-1}{2}\left[\text{log}\left( \frac{x^{'}}{\sigma_x} \right)^2+ \text{log}\left( \frac{y^{'}}{\sigma_y}\right)^2\right] \right\}\text{sin}(2\pi fx^{'}),
\end{aligned}
\vspace{-5pt}
\end{equation}
where
\begin{equation}
\begin{aligned}
&x^{'} = x{\rm cos}\theta + y{\rm{sin}}\theta;\\
&y^{'} = y{\rm cos}\theta - x{\rm{sin}}\theta,
\end{aligned}
\end{equation}
% where $f$ is the frequency of the sinusoidal wave $(x^\prime,y^\prime)$, and $\theta$ is the rotation angle of the sinusoidal wave, $\sigma_x$ and $\sigma_y$ are the standard deviations of the Gaussian function in the $x$-direction and $y$-direction, respectively. 
where $f$ and $\theta$ denote the frequency and the rotation angle of $(x^\prime,y^\prime)$. The standard deviations of the Gaussian envelope in the two directions
% , $x$ and $y$, 
are represented as $\sigma_x$ and $\sigma_y$, respectively.
% $\sigma_x$ and $\sigma_y$ represent the standard deviations of the Gaussian function in the $x$-direction and $y$-direction.
More specifically, in this work, the rotation angle is set to $0$ and $\pi / 2$ to extract horizontal and vertical edge features, respectively, and the frequency $f$ is empirically set to 2.5. 
Furthermore, the HDR images can provide abundant texture information in the bright areas, which enables it to provide the better visual experience compared with the LDR image. Therefore, a spatial mask is applied on the extracted local features, focusing on the high luminance region of HDR images.
% local edge features are extracted from the generated global edge features via a spatial mask, focusing on the high luminance region of HDR images. 
The designed Gaussian function is applied to obtain the mask, 
\begin{equation}
M_g = 1 + \frac{1}{2\pi\sigma}{\rm {exp}}\left(\frac{-(L_r)-\mu)^2}{2\sigma^2}\right),
\end{equation}
where $\sigma$ and $\mu$ are empirically set to 0.2 and 250.
% in this work. 
% $L_r$ is the luminance of HDR image. 
It is worth mentioning that all feature extraction operations in this work are performed on the luminance of HDR images unless otherwise stated.

\subsection{Butterworth Filter-based Global Feature Extraction}
Over the past few decades, frequency features have been widely adopted to extract the local texture or edge information of the nature images~\cite{yang2008discrete, ni2018gabor}. 
However, the direct difference in the frequency domain between the reference and distorted images has not been extensively discussed. 
% Furthermore, as the spatial frequency increases, the contrast sensitivity of the human eye first increases and then decreases~\cite{}, which means there exists a frequency interval that the HVS is highly sensitive to. 
Furthermore, the contrast sensitivity of the human eye first increases and then decreases with the increase of spatial frequency~\cite{barten2003formula}, which implies that there is a frequency interval where HVS is highly sensitive.
Motivated by this observation, this paper adopts the Butterworth filter to simulate the \textit{contrast sensitivity function} (CSF) to directly extract the global feature from the frequency spectrum of the image. 
Given a 2D image, each pixel of its frequency representation is computed from all the pixels in its spatial domain. 
Therefore, the frequency representation of an image can be regarded as a global feature of the image. 
Taking the reference HDR image as an example, the corresponding 2D \textit{Discrete Fourier transform} (DFT) is performed as:
\begin{equation}
F_r(u,v) = \sum\limits_{x=0}^{M-1}\sum\limits_{y=0}^{N-1}L_r(x,y)e^{-j2\pi(ux/M+vy/N)},
\end{equation}
where $M$ and $N$ are the size of the image, $(u,v)$ denotes the pixel coordinate in the frequency spectrum. 
After shifting the low frequencies to the middle of the frequency spectrum, the $\rm log$ operation is applied to compress the values for better representation. 
Subsequently, a bandpass Butterworth filter is designed to provide higher weights to the special frequency interval,
\begin{equation}
M_b = \left(1-\frac{1}{1+(D_1/D)^{2n_1}}\right)\cdot\left(\frac{1}{1+(D_2/D)^{2n_2}}\right),
\end{equation}
where $D$ is the Euclidean distance of the 2D grid. 
In this work, the cut-off frequency and order value $D_1$, $D_2$, $n_1$, and $n_2$ are empirically set to 400, 100, 4 and 2, respectively. 
The masked frequency map is presented as,
\begin{equation}
% B_r^{f}(x,y) = {\rm{log}}({{\rm{abs}}(F_r(u,v))}+1)\cdot M_b
B_r^{f}(x,y) = {\rm{log}}({|F_r(u,v)|}+1)\cdot M_b.
\end{equation}

Moreover, by separating the frequency spectrum into real and imaginary part, i.e., $R_r=Real(F_r)$ and $I_r=Imag(F_r)$, the phase map can be obtained as $B_r^{p}= {\rm{tan}^{-1}(I_r/R_r)}$.
In the same way, the frequency map $B_d^{f}$ and phase map $B_d^{p}$ of the distorted HDR image can be obtained. 
Therefore, the Butterworth filter-based global feature is composed of the frequency map and phase map, which is given by, 
% , which can be presented as:
%That is,
\begin{equation}
\begin{aligned}
B_r(x,y) = \{B_r^f(x,y), B_r^p(x,y)\};\\
B_d(x,y) = \{B_d^f(x,y), B_d^p(x,y)\}.
\end{aligned}
\end{equation}

% \vspace{-10pt}
\subsection{Feature Similarity Measurements}
Since the generated local and global frequency features are represented in different domains, the feature similarity measurements are conducted on the spatial and frequency domains, respectively. 
% For the local frequency features, the similarity map can be measured as,
The similarity map of the local frequency features is calculated as follows:
\begin{equation}
\begin{aligned}
% S_l^l(x,y) = \frac{2G_r^l(x,y)\cdot G_d^l(x,y)+T_0}{G_r^l(x,y)^2+G_d^l(x,y)^2+T_0};\\
% S_l^g(x,y) = \frac{2G_r^g(x,y)\cdot G_d^g(x,y)+T_0}{G_r^g(x,y)^2+G_d^g(x,y)^2+T_0},
S_L(x,y) = \frac{2G_r(x,y)\cdot G_d(x,y)+T_0}{G_r(x,y)^2+G_d(x,y)^2+T_0},
\end{aligned}
\end{equation}
where the $T_0$ is a positive constant to prevent the exception that the denominator equals to zero.
% prevent the denominator converges to zero. 
% The local similarity map of the edge feature is obtained by combining the $S_l^l(x,y)$ and $S_l^g(x,y)$:
% \begin{equation}
% S_L(x,y) = [S_l^l(x,y)]^\alpha\cdot[S_l^g(x,y)]^{(1-\alpha)},
% \end{equation}
% where $\alpha$ is a weighting parameter to adjust the relative importance of $S_l^l(x,y)$ and $S_l^g(x,y)$.

For the global feature, the similarity of the frequency map and phase map can be generated as,
\begin{equation}
\begin{aligned}
S_g^f(x,y) = \frac{2B_r^f(x,y)\cdot B_d^f(x,y)+T_1}{B_r^f(x,y)^2+B_d^f(x,y)^2+T_1};\\
S_g^p(x,y) = \frac{2B_r^p(x,y)\cdot B_d^p(x,y)+T_2}{B_r^p(x,y)^2+B_d^p(x,y)^2+T_2},
\end{aligned}
\end{equation}
where the $T_1$ and $T_2$ are the positive constants,
% to guarantee the numerical stability.
in analogous to $T_0$.
The final similarity map of the global frequency feature is obtained as,
\begin{equation}
S_G(x,y) = [S_g^f(x,y)]^\alpha\cdot[S_g^p(x,y)]^{(1-\alpha)},
\end{equation}
where $\alpha$ is a positive constant used for weighting control of $S_g^f(x,y)$ and $S_g^p(x,y)$.
% that are used to control the weighting of $S_g^f(x,y)$ and $S_g^p(x,y)$. 
In this work, $T_0$, $T_1$, $T_2$, and $\alpha$ are empirically set as 0.014, 8, 1, and 0.5, respectively.

% \vspace{-10pt}
\subsection{Feature Pooling}
In feature maps generated from HDR images using the Gabor filter and Butterworth filter, the larger pixel value implies that the HVS is more sensitive and pays more attention to it. 
Therefore, the weighted maps for the local and global frequency feature maps can be generated as:
\begin{equation}
\begin{aligned}
W_G(x,y) = {\rm{max}}\{|G_r(x,y)|,|G_d(x,y)|\};\\
W_L(x,y) = {\rm{max}}\{|B_r^f(x,y)|,|B_d^f(x,y)|\}.
\end{aligned}
\end{equation}

Therefore, the local frequency similarity score and global frequency similarity score can be calculated as the weighted average over all the pixel locations $(x, y)$ on the corresponding similarity maps as:
% the local and global quality score can be calculated as,
\begin{equation}
\begin{aligned}
% Q_G(x,y) = \frac{\sum\limits_{(x,y)}W_G(x,y)\cdot S_G(x, y)}{\sum\limits_{(x,y)}W_G(x,y)};\\
% Q_L(x,y) = \frac{\sum\limits_{(x,y)}W_L(x,y)\cdot S_L(x, y)}{\sum\limits_{(x,y)}W_L(x,y)}.
Q_G(x,y) = \sum\limits_{(x,y)}W_G(x,y)\cdot S_G(x, y) / \sum\limits_{(x,y)}W_G(x,y);\\
Q_L(x,y) = \sum\limits_{(x,y)}W_L(x,y)\cdot S_L(x, y) / \sum\limits_{(x,y)}W_L(x,y).
\end{aligned}
\end{equation}
% The final HDR IQA score is obtained by combining the local and global quality scores,

The final quality score is obtained by combining the local similarity score and global frequency similarity score,
\begin{equation}
Q_{LGFM} = Q_G(x,y)\cdot Q_L(x,y).
\end{equation}

\section{EXPERIMENTAL RESULTS}
\label{sec:results}

\subsection{HDR Dataset and Evaluation Protocols}

In this section, four publicly available datasets are used for performance evaluation~\cite{valenzise2014performance}~\cite{zerman2017extensive}~\cite{narwaria2014impact}~\cite{mikhailiuk2020upiq}. 
The first three datasets are constructed by subjective experiments, while the fourth dataset consists of two existing datasets using a specifically designed algorithm to align subjective scores~\cite{narwaria2014impact}~\cite{korshunov2015subjective}. 

\begin{table*}[htbp]
\footnotesize
% \normalsize
\renewcommand{\arraystretch}{1.1}
\tabcolsep0.2cm
  \centering
  \caption{\emph{SROCC}, \emph{KROCC} and \emph{RMSE} Comparison of various IQA models on four widely used benchmarks.}
    \begin{tabular} {cp{0.6cm}<{\raggedright}|p{0.2cm}<{\centering}p{0.6cm}<{\centering}p{0.6cm}<{\centering}p{0.6cm}<{\centering}p{0.6cm}<{\centering}|p{0.6cm}<{\centering}p{0.6cm}<{\centering}p{0.6cm}<{\centering}p{0.6cm}<{\centering}p{0.6cm}<{\centering}|p{0.6cm}<{\centering}p{0.6cm}<{\centering}p{0.6cm}<{\centering}p{0.6cm}<{\centering}p{0.6cm}<{\centering}p{0.6cm}<{\centering}}
    % \toprule
    \hline
    \hline
    \multicolumn{2}{c|}{Criteria} & \multicolumn{5}{c|}{SROCC}    & \multicolumn{5}{c|}{KROCC}    & \multicolumn{5}{c}{RMSE} \\
    % \midrule
    \hline
     \multicolumn{2}{c|}{Dataset} & 
     \multicolumn{1}{p{2.2em}}{D-V} & 
     \multicolumn{1}{p{2.2em}}{D-Z}& 
     \multicolumn{1}{p{2.2em}}{D-N} & 
     \multicolumn{1}{p{2.2em}}{UPIQ} & 
     \multicolumn{1}{p{2.4em}|}{Avg.} & 
     \multicolumn{1}{p{2.2em}}{D-V} & 
     \multicolumn{1}{p{2.2em}}{D-Z}& 
     \multicolumn{1}{p{2.2em}}{D-N} & 
     \multicolumn{1}{p{2.2em}}{UPIQ} & 
     \multicolumn{1}{p{2.4em}|}{Avg.} & 
     \multicolumn{1}{p{2.2em}}{D-V} & 
     \multicolumn{1}{p{2.2em}}{D-Z}& 
     \multicolumn{1}{p{2.2em}}{D-N} & 
     \multicolumn{1}{p{2.2em}}{UPIQ} &
     \multicolumn{1}{p{2.6em}}{Avg.}\\
    % \midrule
    % \midrule
    \hline
       \multicolumn{2}{c|}{HDR-VQM} 
     & 0.8965  & 0.7611  & 0.7248  & \textbf{0.8888 } & 0.8178  & 0.7031  & 0.5621  & 0.5649  & \textbf{0.7002 } & 0.6326  & 12.2101  & 18.8356  & 0.8189  & 0.3819  & \textbf{8.0616 }\\
    % \midrule
    \hline
    \multicolumn{2}{c|}{HDR-VDP-3} 
    & 0.9132  & 0.7558  & \textbf{0.8467} & 0.8214  & 0.8343 & \textbf{0.7392} & 0.5698  & \textbf{0.6718} & 0.6421  & 0.6557 & \textcolor{blue}{\textbf{10.7257}} & 19.6734  & \textcolor{blue}{\textbf{0.5470}} & 0.4089 & 7.8388 \\
    % \midrule
    \hline
    \multirow{2}[2]{*}{SSIM} 
    & PU    & 0.8699  & 0.7827  & 0.7057  & 0.7369  & 0.7738  & 0.6719  & 0.5787  & 0.5332  & 0.5488  & 0.5832  & 14.0147  & 18.7897  & 0.8298  & 0.5258  & 8.5400  \\
          & PQ    & 0.8736  & 0.7682  & 0.6508  & 0.7315  & 0.7560  & 0.6850  & 0.5783  & 0.4870  & 0.5486  & 0.5747  & 14.5140  & 19.4345  & 0.8665  & 0.5132  & 8.8321\\
    % \midrule
    \hline
    \multirow{2}[2]{*}{MSSIM} 
    & PU    & 0.9009  & 0.8004  & 0.8377  & 0.8063  & 0.8363  & 0.7261  & 0.6098  & 0.6622  & 0.6113  & 0.6524  & 12.1315  & 29.1735  & 0.7774  & 0.5162  & 10.6497  \\
          & PQ    & 0.8994  & 0.8417  & 0.8299  & 0.8118  & 0.8457  & 0.7113  & 0.6458  & 0.6529  & 0.6197  & 0.6574  & 11.8489  & 15.8932  & 0.7801  & 0.5023  & 7.2561  \\
    % \midrule
    \hline
    \multirow{2}[2]{*}{PSNR} 
    & PU    & 0.6672  & 0.6591  & 0.6131  & 0.6311  & 0.6426  & 0.4797  & 0.4760  & 0.4517  & 0.4582  & 0.4664  & 22.6432  & 21.8065  & 0.8581  & 0.5208  & 11.4572  \\
          & PQ    & 0.6681  & 0.6476  & 0.6198  & 0.6327  & 0.6421  & 0.4797  & 0.4623  & 0.4571  & 0.4588  & 0.4645  & 21.3461  & 20.0826  & 0.8619  & 0.5147  & 10.7013  \\
    % \midrule
    \hline
    \multirow{2}[2]{*}{GMSD} 
    & PU    & 0.8918  & 0.8738  & 0.8080  & 0.7784  & 0.8380  & 0.7064  & 0.6919  & 0.6224  & 0.5824  & 0.6508  & 12.0439  & 13.4912  & 0.6487  & 0.4357  & 6.6549  \\
          & PQ    & 0.8846  & 0.8719  & 0.7979  & 0.7696  & 0.8310  & 0.6900  & 0.6886  & 0.6134  & 0.5735  & 0.6414  & 12.3418  & 13.6594  & 0.6798  & 0.4550  & 6.7840  \\
    % \midrule
    \hline
    \multirow{2}[2]{*}{FSIM} 
    & PU    & 0.9090  & 0.8337  & 0.8323  & 0.7693  & 0.8361  & 0.7376  & 0.6571  & 0.6525  & 0.5702  & 0.6544  & 11.8094  & 15.4822  & 0.7790  & 0.5258  & 7.1491  \\
          & PQ    & 0.9064  & 0.8358  & 0.8213  & 0.7648  & 0.8321  & 0.7392  & 0.6583  & 0.6401  & 0.5669  & 0.6511  & 11.7802  & 15.5461  & 0.8005  & 0.5378  & 7.1662  \\
    % \midrule
    \hline
    \multirow{2}[2]{*}{VIF} 
    & PU    & \textbf{0.9183 } & 0.8284  & 0.6705  & 0.7640  & 0.7953  & 0.7372  & 0.6737  & 0.5067  & 0.5838  & 0.6254  & 11.4667  & 17.3586  & 0.7338  & 0.4233  & 7.4956  \\
          & PQ    & 0.9158  & 0.8283  & 0.6631  & 0.7568  & 0.7910  & 0.7307  & 0.6753  & 0.4977  & 0.5761  & 0.6200  & 11.4650  & 18.2018  & 0.7442  & 0.4309  & 7.7105  \\
    % \midrule
    \hline
    \multirow{2}[2]{*}{ESIM} 
    & PU    & 0.9176  & 0.8866  & 0.7763  & 0.8559  & \textbf{0.8591 } & 0.7392  & 0.7137  & 0.5932  & 0.6762  & \textbf{0.6806 } & 12.4762  & 13.8699  & 0.6487  & 0.3565  & 6.8378  \\
          & PQ    & 0.9139  & 0.8836  & 0.7685  & 0.8529  & 0.8547  & 0.7392  & 0.7088  & 0.5863  & 0.6720  & 0.6766  & 12.5785  & 13.7950  & 0.6480  & \textbf{0.3518 } & 6.8433  \\
    % \midrule
    \hline
    \multirow{2}[2]{*}{GFM} 
    & PU    & 0.9052  & \textbf{0.9154} & 0.7182  & 0.7848 & 0.8309 & 0.7162  & \textbf{0.7485} & 0.5400  & 0.5890   & 0.6484 & 11.7361  & \textcolor{blue}{\textbf{10.8275}} & 0.7617  & 0.4494  & \textbf{5.9437}\\
          & PQ    & 0.8984  & 0.9088  & 0.7119  & 0.7693 & 0.8221 & 0.7113  & 0.7359  & 0.5380  & 0.5738 & 0.6398 & 11.9960  & 17.8278  & 0.7922  & 0.4767  & 7.7732 \\
    % \midrule
    \hline
    \multirow{2}[2]{*}{LGFM} 
    & PU  & \textcolor{red}{\textbf{0.9200}} & \textcolor{red}{\textbf{0.9322}} & \textcolor{blue}{\textbf{0.8539}} & \textcolor{red}{\textbf{0.9032}} & \textcolor{red}{\textbf{0.9023}} & \textcolor{red}{\textbf{0.7491}} & \textcolor{red}{\textbf{0.7707}} & \textcolor{blue}{\textbf{0.6824}} & \textcolor{red}{\textbf{0.7236}} & \textcolor{red}{\textbf{0.7315}} & \textcolor{red}{\textbf{10.6462}} & \textcolor{red}{\textbf{10.5293}} & \textcolor{red}{\textbf{0.5392}} & \textcolor{blue}{\textbf{0.3097}} & \textcolor{red}{\textbf{5.5061}} \\
          & PQ    & \textcolor{blue}{\textbf{0.9195}} & \textcolor{blue}{\textbf{0.9207}} & \textcolor{red}{\textbf{0.8611}} & \textcolor{blue}{\textbf{0.8928}} & \textcolor{blue}{\textbf{0.8985}} & \textcolor{blue}{\textbf{0.7458}} & \textcolor{blue}{\textbf{0.7634}} & \textcolor{red}{\textbf{0.6833}} & \textcolor{blue}{\textbf{0.7070}} & \textcolor{blue}{\textbf{0.7249}} & \textbf{11.4299} & \textbf{10.9682} & \textbf{0.5523} & \textcolor{red}{\textbf{0.3072}} & \textcolor{blue}{\textbf{5.8144}} \\
    % \bottomrule
    \hline
    \hline
    \end{tabular}%
  \label{tab:performance}%
     \vspace{-10pt}
\end{table*}%

As suggested in the VQEG HDTV test~\cite{video2000final}~\cite{sheikh2006statistical}, a logistic regression function is applied to map the predicted objective scores to a common scale,
% the subjective scores.
% to perform the nonlinear mapping from the predicted objective scores to the subjective scores:
\begin{equation}
S_i = \gamma_1\left(\frac{1}{2}-\frac{1}{1+e^{\gamma_2(q_i-\gamma_3)}}\right) + \gamma_4q_i + \gamma_5,
\end{equation}
% where $q_i$ is the quality score of the $i$-th image predicted by the IQA model, $S_i$ is the corresponding subjective score. 
where $q_i$ and $S_i$ denote the generated quality score of the $i$-th image from the IQA model and the corresponding mapped score.
$\gamma_1$, $\gamma_2$, $\gamma_3$, $\gamma_4$, and $\gamma_5$ are the regression parameters determined by minimizing the sum of squared differences between the predicted objective score $S_i$ and corresponding ground truth score (i.e., MOS/DMOS). 
% three commonly-used criteria, namely, 
Subsequently, the \textit{Spearman rank order correlations coefficient} (SROCC), \textit{Kendall rank order correlation coefficient}, and \textit{root mean square error} (RMSE) are adopted to evaluate the performance of various HDR IQA metrics. 
Note that higher values of SROCC and KROCC represent a stronger correlation, while lower values of RMES indicate smaller differences.

\begin{table}[htbp]
\footnotesize
% \normalsize
\renewcommand{\arraystretch}{1.0}
\tabcolsep0.15cm
  \centering
  \caption{Ablation study to explore the role of the local and global frequency features, as well as the masks $M_g$ and $M_b$.}
    \begin{tabular}{ccccccc}
    % \toprule
    \hline
    \hline
          & {Dataset} & 
          {L}& 
          {L w/o $M_g$} & 
          {G} & 
          {G w/o $M_b$} & 
          {LGFM} \\
    % \midrule    
    \hline
    \multirow{4}[2]{0em}{\rotatebox{90}{SROCC}} & D-V   & 0.9139  & 0.9137  & \textcolor{blue}{\textbf{0.9163}}  & 0.9131  & \textcolor{red}{\textbf{0.9200}}  \\
          & D-Z   & \textcolor{blue}{\textbf{0.9249}}  & \textbf{0.9244}  & 0.9234  & 0.9128  & \textcolor{red}{\textbf{0.9322}}  \\
          & D-N   & 0.8355  & \textbf{0.8382}  & \textcolor{blue}{\textbf{0.8484}}  & 0.8346  & \textcolor{red}{\textbf{0.8539}}  \\
          & UPIQ   & \textbf{0.8717}  & 0.8707  & \textcolor{blue}{\textbf{0.8982}}  & 0.8516  & \textcolor{red}{\textbf{0.9032}}  \\
    % \midrule
    \hline
    \multirow{4}[2]{0em}{\rotatebox{90}{KROCC}} & D-V   & \textcolor{blue}{\textbf{0.7409}}  & \textbf{0.7392}  & 0.7376  & 0.7310  & \textcolor{red}{\textbf{0.7491}}  \\
          & D-Z   & \textcolor{blue}{\textbf{0.7574}}  & \textbf{0.7561}  & 0.7537  & 0.7319  & \textcolor{red}{\textbf{0.7707}}  \\
          & D-N   & 0.6575  & \textbf{0.6593}  & \textcolor{blue}{\textbf{0.6768}}  & 0.6577  & \textcolor{red}{\textbf{0.6824}}  \\
          & UPIQ   & \textbf{0.6831}  & 0.6816  & \textcolor{blue}{\textbf{0.7230}}  & 0.6617  & \textcolor{red}{\textbf{0.7236}}  \\
    % \midrule
    \hline
    \multirow{4}[2]{0em}{\rotatebox{90}{RMSE}} & D-V   & \textcolor{blue}{\textbf{11.0598}}  & \textbf{11.1183}  & 11.5351  & 14.3446  & \textcolor{red}{\textbf{10.6462}}  \\
          & D-Z   & \textcolor{blue}{\textbf{10.7154}}  &\textbf{10.7952}  & 11.3910  & 13.7609  & \textcolor{red}{\textbf{10.5293}}  \\
          & D-N   & 0.5556  & \textcolor{blue}{\textbf{0.5535}}  & \textbf{0.5431}  & 0.6718  & \textcolor{red}{\textbf{0.5392}}  \\
          & UPIQ   & \textbf{0.3210}  & 0.3227  & \textcolor{blue}{\textbf{0.3114}}  & 0.3945  & \textcolor{red}{\textbf{0.3097}}  \\
    % \bottomrule
    \hline
    \hline
    \end{tabular}%
  \label{tab:ablation}%
     \vspace{-10pt}
\end{table}%

\vspace{-10pt}
\subsection{Performance Comparison}
To illustrate the superiority of the proposed LGFM, several classic and state-of-the-art IQA metrics are adopted for comparison, including HDR-VDP-3~\cite{narwaria2015hdr}, HDR-VQM~\cite{narwaria2015hdrvqm}, PSNR, SSIM~\cite{wang2004image}, FSIM~\cite{zhang2011fsim}, VIF~\cite{sheikh2005information}, GMSD~\cite{xue2013gradient}, ESIM~\cite{ni2017esim}, GFM~\cite{ni2018gabor}, and MSSIM~\cite{wang2003multiscale}, where the first two algorithms are specifically designed for HDR images, while the other metrics are for LDR images. 
Therefore, HDR images are first converted into perceptual space by PU or PQ coding before applying the LDR IQA metrics. 
Since the parameters in part of the IQA models (i.e., display and distance parameters in HDR-VDP-3 and HDR-VQM) would affect the final results, we use the default settings during the experiments.

% \subsubsection{Overall performance evaluation}
% Table~\ref{tab:performance} presents the performance comparison of various IQA models on four datasets, respectively. 
% The first-ranked, the second-ranked, and the third-ranked performance of each measurement criterion (i.e., SROCC, KROCC, or RMSE) are highlighted in bold with red, blue, and black, respectively. 
Table~\ref{tab:performance} presents the performance comparison of various IQA models on four datasets, where the highlighted in bold with red, blue, and black represent the first, second, and third-ranked performance of the measurement criterions.
% (i.e., SROCC, KROCC, or RMSE).
Compared with the state-of-the-art IQA metrics, our proposed LGFM model yields the best overall performance in terms of SROCC, KROCC, and MSE on four datasets. 
Besides, the HDR-VDP-3, HDR-VQM, and PU-GFM also achieved relatively promising results. 
Moreover, there is an interesting observation that most of the highlighted values are from PU coding, which indicates that PU encoding performs higher consistency with the HVS compared with the PQ coding.

\vspace{-10pt}
\subsection{Ablation Study}
This subsection verifies the effectiveness of each component in the proposed LGFM model, including the Gabor filter-based local feature, Butterworth filter-based global feature, as well as the proposed two masks $M_g$ and $M_b$. 
% In this section, the effectiveness of each component in the proposed LGFM model is verified, 
More specifically, four LGFM variants are conducted as follows: 
1) {\rm L}: only Gabor filter-based local frequency feature is used for evaluation. 
2) {\rm L w/o} $M_g$: removing the weighting mask $M_g$ in $\rm L$. 
3) \rm G: only Butterworth filter-based global frequency feature is used for evaluation. 
4) {\rm G w/o} $M_b$: removing the weighting mask $M_b$ in $\rm G$. 
% in $\rm G$, replacing the $M_b$ by a metric with the same size, where each pixel is equal to 1. 
It is worth mentioning that PU coding is adopted in our model. 
% In Table~\ref{tab:ablation}, the highlighted in bold with red, blue, and black represent the first, second, and third-ranked performance of the measurement criterions (i.e., SROCC, KROCC, or RMSE), respectively. 
As shown in Table~\ref{tab:ablation}, both the $M_g$ and $M_b$ 
% weighting maps 
contribute to feature extraction in $\rm L$ and $\rm G$, respectively, which illustrate the effectiveness of the over-exposed regions and the frequency interval. 
Furthermore, the combination of local and global frequency features outperforms using either one alone, indicating that the proposed two feature maps achieve complementarity between local and global frequency features.
% Moreover, the performance of the combination of the local and global frequency features is superior to using either one alone, which demonstrates that the proposed two feature maps achieve complementarity between local and global frequency features.

\section{CONCLUSIONS}
\label{sec:conclusion}
% In this paper, a novel \textit{local and global frequency feature-based model} (LGFM) is proposed for \textit{high dynamic range} (HDR) image quality assessment.
This paper proposes the LGFM model, a novel image quality assessment algorithm for HDR images.
The key contribution of our model lies in the combination of the local and global frequency features. 
More specifically, the local feature map is extracted by the Gabor filter to measure the structure similarity, while the global feature map is obtained by simulating the \textit{contrast sensitivity function} with the Butterworth filter to detect the frequency interval similarity. Subsequently, the feature pooling strategy is adopted to generate the quality scores based on the local and global similarity maps, leading to the final quality score by combing them. 
Extensive experiments demonstrate that each component in the proposed LGFM contributes to the final results. 
Moreover, the proposed model provides higher consistency with the HVS and outperforms other state-of-the-art IQA models.

\bibliographystyle{IEEEtran}
\bibliography{egbib}

\end{document}